\documentclass[11pt]{article}

\usepackage[preprint]{acl}

\usepackage{times}
\usepackage{latexsym}
\usepackage[T1]{fontenc}
\usepackage[utf8]{inputenc}
\usepackage{microtype}
\usepackage{inconsolata}
\usepackage{graphicx}
\usepackage{booktabs}
\usepackage{amsmath}
\usepackage{amssymb}
\usepackage{xcolor}
\usepackage{enumitem}
\usepackage{subcaption}
\usepackage{xspace}
\usepackage{pifont}
\usepackage{multirow}
\usepackage{diagbox}
\usepackage{url}
\usepackage{balance}
\usepackage{placeins}
\usepackage{listings}
\usepackage{tcolorbox}
\tcbuselibrary{listings,skins,breakable}

\definecolor{promptbg}{RGB}{248,248,248}
\definecolor{promptframe}{RGB}{180,180,180}
\definecolor{prompttitle}{RGB}{60,60,60}
\newtcblisting{promptbox}[1]{
  enhanced,
  colback=promptbg,
  colframe=promptframe,
  coltitle=prompttitle,
  fonttitle=\bfseries\small,
  title={#1},
  listing only,
  left=4pt, right=4pt, top=2pt, bottom=2pt,
  boxrule=0.5pt,
  arc=2pt,
  listing options={
    basicstyle=\ttfamily\fontsize{7}{8.5}\selectfont,
    breaklines=true,
    breakatwhitespace=false,
    columns=fullflexible,
    keepspaces=true,
    showstringspaces=false,
    tabsize=2,
    aboveskip=0pt,
    belowskip=0pt,
  },
}

\newcommand{\cmark}{\ding{51}}
\newcommand{\xmark}{\ding{55}}
\newcommand{\ours}{\textsc{Web\kern0.03em X\kern-0.02em Skill}\xspace}

\title{\ours: Skill Learning for Autonomous Web Agents}

\author{
Zhaoyang Wang$^{1}$, Qianhui Wu$^{2,*}$, Xuchao Zhang$^{2}$, Chaoyun Zhang$^{2}$, Wenlin Yao$^{2}$,\\
\scalebox{0.96}{\textbf{Fazle Elahi Faisal$^{2}$, Baolin Peng$^{2}$, Si Qin$^{2}$, Suman Nath$^{2}$, Qingwei Lin$^{2}$, Chetan Bansal$^{2}$,}}\\
\textbf{Dongmei Zhang$^{2}$, Saravan Rajmohan$^{2}$, Jianfeng Gao$^{2}$, Huaxiu Yao$^{1,*}$}\\[5pt]
$^{1}$University of North Carolina at Chapel Hill \qquad $^{2}$Microsoft\\
\texttt{\{zhaoyang,huaxiu\}@cs.unc.edu} \quad
\texttt{\{qianhuiwu,xuchaozhang\}@microsoft.com}
}

\begin{document}
\maketitle
\begingroup
\renewcommand{\thefootnote}{\fnsymbol{footnote}}
\footnotetext[1]{Corresponding authors.}
\endgroup

\begin{abstract}
  Autonomous web agents powered by large language models (LLMs)  remain brittle on long-horizon browser workflows. A key bottleneck is a grounding gap in existing skill formulations: textual workflow skills provide natural language guidance but cannot be directly executed, while code-based skills execute without giving the agent step-level guidance for adaptation or recovery. We introduce \ours, a framework that bridges this gap with executable skills, each pairing a parameterized action program with step-level natural-language guidance. \ours operates in three stages: skill extraction mines reusable action subsequences from readily available synthetic agent trajectories and abstracts them into parameterized skills, skill organization indexes them into a URL-based graph for context-aware retrieval, and skill deployment exposes two complementary modes, grounded mode for fully automated execution and guided mode where skills serve as step-by-step instructions the agent follows with its native planning. \ours demonstrates consistent improvements on WebArena, WebVoyager, and Online-Mind2Web. We further find that better skill deployment mode depends on a model's plan and execution capability. The code is available at \url{https://github.com/aiming-lab/WebXSkill}.
\end{abstract}

\section{Introduction}
\label{sec:intro}

Large language models (LLMs) have enabled autonomous web agents that interact with real websites through browser actions such as clicking, typing, and navigating~\citep{agashe2025agent,murty2025nnetnavunsupervisedlearningbrowser,ning2025surveywebagentsnextgenerationai,webarena,he2024openwebvoyager}. Yet these agents remain brittle on complex multi-page workflows because they rarely retain and reuse procedural interaction knowledge, forcing them to re-plan long action sequences from scratch even for recurring routines~\citep{onlinemind2web,li2025websailor}. For example, when an agent completes a complex task, the procedural guidance embedded in the trajectory is often discarded, and the next time the agent encounters a similar task it must re-derive the entire action sequence, increasing the risk of planning and execution errors.

\begin{figure}[t]
  \centering
  \includegraphics[width=0.88\columnwidth]{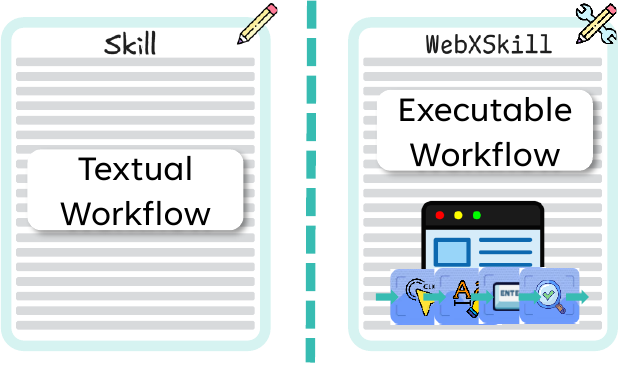}
  \caption{\ours equips web agents with executable skills that pair an action program with step-level natural-language guidance for planning assistance.}
  \label{fig:teaser}
  \vspace{-3pt}
\end{figure}

Recent work has begun to address this gap through skills, reusable knowledge units that encode common interaction patterns~\citep{sodhi2023step,agent-workflow-memory,walt,skillweaver,asi}. However, existing skill formulations for web agents suffer from a grounding gap. Workflow-based approaches such as AWM~\citep{agent-workflow-memory} represent skills as natural language instructions (e.g., ``search for the product, then add it to cart'') that guide planning but cannot be directly executed, requiring the agent to translate each instruction into concrete browser actions and reintroducing grounding errors. Code-based approaches like SkillWeaver~\citep{skillweaver} and WALT~\citep{walt} ground skills in executable code or action scripts, but deploy them as opaque black-box calls without step-level guidance, making the agent unable to understand, adapt, or recover when execution fails mid-way. Note that detailed code scripts are often not provided to the agent, making it impossible for the agent to learn from the skill's internal logic or to flexibly adapt to actual situations. 

In this paper, we propose \ours, a framework that bridges the grounding gap by introducing \emph{executable skills} that pair parameterized action programs with step-level natural-language guidance, as shown in Figure~\ref{fig:teaser}. Each skill carries both a concrete sequence of browser operations and semantic annotations (name, description, typed parameters, and per-step guidance), making it simultaneously executable by the runtime and readable by the agent. This dual nature enables two complementary deployment modes: a \emph{grounded mode} where the agent invokes a skill as an atomic tool call and the runtime auto-executes the underlying action sequence, and a \emph{guided mode} where skills are surfaced as step-by-step instructions the agent follows using its native browser actions, preserving autonomy for adaptation. In contrast to prior work that hides the internal logic of skills, our two modes offer the agent readable details of the selected skill, providing agent autonomy to adapt in unexpected states.
\ours obtains such skills by:
(1) skill extraction, which mines reusable action subsequences from abundant synthetic agent trajectories and abstracts them into parameterized skills via LLM-based generalization;
(2) skill organization, which indexes skills into a URL-based skill graph that maps web pages to their applicable skills, enabling context-aware retrieval; and
(3) skill deployment, which selects between grounded and guided execution depending on the task and model capabilities.
Our contributions are:
\begin{itemize}
    \item We identify a grounding gap between textual and code-based skill formulations for web agents, and propose executable skills that pair parameterized action programs with step-level natural-language guidance, transparent to the agent while maintaining efficiency.
    \item We instantiate this as \ours, which extracts skills from low-cost synthetic trajectories (avoiding test-time leakage), organizes them in a skill graph for context-aware retrieval, and deploys them in grounded or guided mode for skill transfer across websites.
    \item \ours improves task success over strong baseline methods on WebArena, WebVoyager, and Online-Mind2Web.  We further show that our deployment mode can be adaptively selected based on model capabilities.
\end{itemize}

\section{Related Work}
\label{sec:related_work}

\paragraph{Web Agents.}
Web agents have rapidly evolved with LLMs and vision-language models that autonomously interact with browsers to complete complex tasks~\citep{nakano2021webgpt,ning2025surveywebagentsnextgenerationai,he2024openwebvoyager,li2025websailor}. Early work established the ReAct paradigm~\citep{yao2023reactsynergizingreasoningacting}, interleaving chain-of-thought reasoning with browser actions, and benchmarks such as Mind2Web~\citep{deng2023mind2webgeneralistagentweb}, WebVoyager~\citep{he2024webvoyager}, and WebArena~\citep{webarena} evaluate agents on live or self-hosted sites. Complementary directions include planning, agent collaboration, and memory~\citep{gao2023palprogramaidedlanguagemodels,hong2024metagpt,wu2024autogen,chhikara2025mem0,yang2026plugmemtaskagnosticpluginmemory}, data synthesis~\citep{sun2025osgenesisautomatingguiagent,pahuja2025explorerscalingexplorationdrivenweb,wang2025adaptingwebagentssynthetic}, and reinforcement learning~\citep{wei2025webagent,lu2025ui,yang2026gui}. These agents operate over low-level browser actions, which makes long-horizon tasks fragile; tool-use and skill learning have emerged to provide reusable abstractions.

\begin{table}[t]
  \centering
  \footnotesize
  \setlength{\tabcolsep}{3pt}
  \renewcommand{\arraystretch}{1.0}
  \begin{tabular}{@{}lcccc@{}}
    \toprule
    \textbf{Method} & \textbf{Exec.} & \textbf{Guid.} & \textbf{Acq.} & \textbf{Ctx.} \\
    \midrule
    AWM~\citep{agent-workflow-memory}            & \xmark & \cmark & Test  & \xmark \\
    SkillWeaver~\citep{skillweaver}    & \cmark & \xmark & Expl. & \xmark \\
    ASI~\citep{asi}            & \cmark & \xmark & Test  & \xmark \\
    WALT~\citep{walt}           & \cmark & \xmark & Expl. & \xmark \\
    \midrule
    \textbf{\ours} & \cmark & \cmark & Traj. & \cmark \\
    \bottomrule
  \end{tabular}
  \caption{Design axes of skill-based web agent methods. Textual skills guide but cannot execute; code-based skills execute but lack step-level guidance.}
  \label{tab:cmp_baseline}
\end{table}

\paragraph{Tool-Use and Skill Learning.}
Hybrid computer-use agents improve efficiency through multi-action prediction and coding-oriented skills~\citep{yang2025ultracua,song2025coact}. Building on this trend, a growing body of work equips web agents with reusable higher-level knowledge~\citep{agent-workflow-memory,zhou2025proposer,skillweaver,asi,walt,sodhi2023step,ical}. Table~\ref{tab:cmp_baseline} summarizes how these methods differ along four axes: executability (\textbf{Exec.}), step-level guidance (\textbf{Guid.}), acquisition strategy (\textbf{Acq.}), and context-aware skill retrieval (\textbf{Ctx.}). Textual workflow methods such as AWM~\citep{agent-workflow-memory}, StepP~\citep{sodhi2023step}, and ICAL~\citep{ical} represent skills as natural language instructions that guide planning but cannot be directly executed, leaving a gap between ``what to do'' and ``how to execute it''. Executable-skill methods close one side of this gap by making skills directly invocable but sacrifice step-level guidance: SkillWeaver~\citep{skillweaver} discovers patterns through autonomous website exploration (\textbf{Expl.}) and compiles them into Python APIs whose internal logic is opaque to the agent; ASI~\citep{asi} induces programmatic skills from successful episodes but acquires them from test-time task trajectories (\textbf{Test}), introducing the risk of evaluation leakage; WALT~\citep{walt} reverse-engineers built-in website functionality into deterministic tools with validated input schemas, again as black-box calls. The internal logic of these code-based skills is hidden from the agent, making it unable to understand, adapt, or recover when execution fails unexpectedly. 
\ours bridges this grounding gap along all axes. Every skill couples an executable program with step-level natural-language guidance, enabling both grounded execution for efficiency and guided step-by-step use for agent autonomy.
Both textual guidance and action programs are transparent and readable to the agent, allowing it to learn from the skill's internal logic and flexibly adapt to actual situations.
Skills are extracted from independently synthesized trajectories (\textbf{Traj.}), avoiding costly exploration and the risk of test-data leakage.
Extracted skills are then organized into a URL-based graph for context-aware retrieval rather than a flat library surfaced regardless of agent's current page state and execution context.

\section{Method}
\label{sec:method}

\paragraph{Problem Setup.}
Given a user task $q$, web agent $\pi$ repeatedly generates browser actions over webpage observations. At each step $t$ the agent receives observation $o_t$ that includes both a textual description of the webpage (e.g., accessibility tree) and a screenshot, and emits an action $a_t$ from a primitive action space $\mathcal{A}_\text{prim}=\{\texttt{click},\texttt{input},\texttt{scroll},\texttt{navigate},\ldots\}$, until the agent signals completion or a step budget is reached. In this web agent setting, the agent often reasons from scratch at every step, with no mechanism to reuse procedural knowledge from prior interactions, even when facing recurring patterns such as searching for a product or filling forms.

\begin{figure*}[t]
  \centering
  \includegraphics[width=1.0\textwidth]{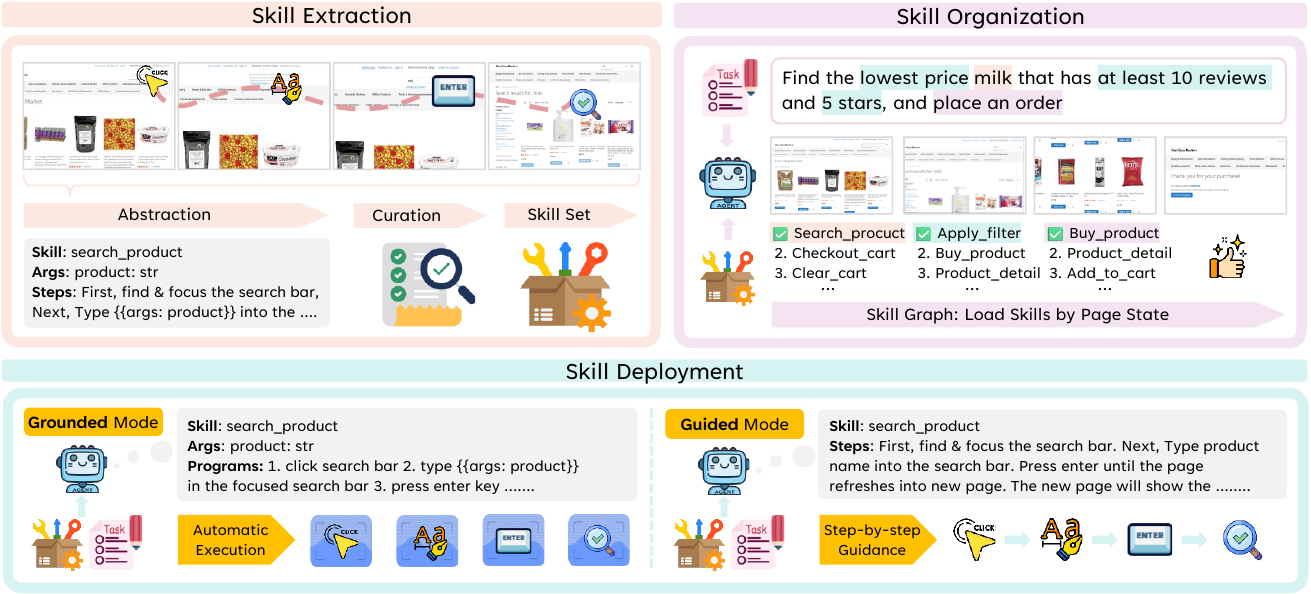}
  \caption{Overview of \ours. (1) \emph{Skill Extraction} abstracts low-level interaction trajectories into reusable skills, followed by curation and verification to improve quality. (2) \emph{Skill Organization} structures skills into a URL-based graph and retrieves state-relevant candidates. (3) \emph{Skill Deployment} supports \emph{grounded mode} (invoke with auto-executed sequence for efficiency) and \emph{guided mode} (surface step-by-step instructions for agent autonomy).}
  \label{fig:pipeline}
\end{figure*}

\paragraph{Overview.}
To mitigate this limitation, we propose \ours, which augments the web agent with a library of executable skills that encode frequently recurring interaction patterns as reusable and readable action programs.
Figure~\ref{fig:pipeline} shows the three stages of our method: skill extraction, skill organization, and skill deployment.

\subsection{Skill Extraction}
\label{sec:extraction}

The goal of skill extraction is to distill agent trajectories into a compact library of parameterized skills. We use readily public available agent trajectories as the data source and exclude any trajectories that are derived from the evaluation benchmarks, avoiding the need for website exploration and the risk of test-data leakage.

\paragraph{Trajectory Abstraction.}
Given a corpus $\mathcal{T}=\{\tau_1,\ldots,\tau_N\}$, where each trajectory records the step-by-step actions an agent took to complete (or attempt) a task, we use an LLM to identify reusable subsequences and abstract them into parameterized skills. 
Specifically, for every $\tau_i$, we construct a structured representation that includes the task description, page URL at each step, the action taken (action type, target element and parameters), and the agent's reasoning. 
The LLM is prompted to: (1) identify action subsequences that represent a coherent, reusable operation (e.g., ``search for a product by keyword''), (2) abstract concrete action values into typed parameters (e.g., replacing a specific search query with a \texttt{query: str} parameter), and (3) annotate each action step with natural language guidance describing its purpose and reasoning.

\paragraph{Skill Curation and Verification.}
To keep the skill library compact without sacrificing coverage, we first use an online deduplication strategy to compare each candidate skill against the existing library before insertion, combining rule-based similarity and embedding-based semantic similarity approaches.
The LLM then decides whether to add, update, or skip the candidate skill. 
We further verify each skill's action sequence by executing it in a test environment and discard any skill that fails to run, which is crucial for effective skill deployment, as studied in our later ablation (Section~\ref{sec:ablation}): reliability is the largest contributor to final accuracy.

\subsection{Skill Organization}
\label{sec:organization}

Extracted skills should be efficiently retrievable at inference time, otherwise massive skill descriptions will overwhelm the agent context.
Unlike existing methods that often treat skills as a flat library or API set to be selected or ranked at every step, we exploit the page-level structure of web browsing: a ``search product'' skill is only applicable on a page with a search bar, not on a checkout page. 

\paragraph{Skill Graph.}
We organize skills into a graph $\mathcal{G}=\{(u_j,\mathcal{S}_j)\}_{j=1}^M$, where $u_j$ is a generalized URL pattern (e.g., \texttt{shopping/catalogsearch/*}) and $\mathcal{S}_j$ contains the skills applicable to matching pages. URL routes are inexpensive signals of page identity and function, while variable path segments can be wildcarded. Skills sharing the same generalized URL are grouped into the same node.
Using this graph will not block skill transfer across sites, since we can still use basic semantic embedding retrieval, as we studied in Section~\ref{sec:results}.

\paragraph{Context-Aware Retrieval.}
At inference time, the agent can efficiently match against graph nodes using the page URL: all matched nodes are retrieved and their associated skills are surfaced as candidates for current step.
Since every skill is bound to specific browser elements, we further filter by checking element presence on the page, ensuring the surfaced skills are executable.
For skill transfer across sites, we use embedding-based retrieval over skill descriptions and guidance to retrieve relevant skills for current task and page context.

\subsection{Skill Deployment}
\label{sec:deployment}
As discussed in Section~\ref{sec:related_work}, existing executable-skill methods deploy skills as black-box calls where the agent has no visibility into the internal logic of the skill, making it unable to adapt when execution fails unexpectedly.
\ours instead exposes both step-level guidance and code-based action sequences to the agent, enabling it to understand the skill logic and flexibly adapt to unexpected states or recover from execution errors. 
This motivates to decouple the ``what'' (the skill's textual guidance and code program) from the ``how'' (the execution strategy): \emph{grounded mode} for automated execution and \emph{guided mode} for agent-driven execution.

\paragraph{Grounded Mode.}
In grounded mode, each skill is exposed as a callable tool in the agent's action space: $\mathcal{A} = \mathcal{A}_\text{prim} \cup \mathcal{A}_\text{skill}$.
The accompanying natural-language guidance remains visible to the agent as a planning aid. 
When the agent invokes a skill (e.g., \texttt{search\_product(query="laptop")}), the runtime executes the corresponding action sequence by matching referenced elements against the current DOM and dispatching the low-level actions in order. 
This mode maximizes efficiency by compressing multi-step procedures into a single call, but it places greater demands on the agent's reasoning and error recovery: execution can be interrupted by unexpected page changes, and the skill may not always align with the agent's intent.

\paragraph{Guided Mode.}
For agents with weaker reasoning capabilities or skills must transfer across sites, guided mode surfaces skills as high-level guidance that the agent follows step-by-step using its own actions, so it can adapt when states differ from expectation.
When activating a skill, the agent receives step-level textual and code guidance (e.g., ``to search a product, first click the search input field, then type the query and press Enter''), while the automated action execution is disabled.
If a step fails due to a changed page layout, the agent can recognize and re-plan toward the same sub-goal through alternative steps.

The two modes can be chosen per agent backbone and scenarios. Grounded mode is more efficient for stronger models that can reliably execute skills and recover from unexpected states; guided mode offers more robustness for weaker models that benefit from explicit procedural guidance. 

\section{Experiments}

\subsection{Experimental Setup}
\label{sec:setup}

\paragraph{Benchmarks.}
We evaluate on three popular web agent benchmarks. WebArena~\citep{webarena} provides five self-hosted websites; following recent work~\citep{liu2025scalecua,yang2026gui}, we use its cleaned 154 task subset. 
Though it is costly for web agent evaluation on live websites due to internet connectivity and reCAPTCHA issues, we also evaluate on two live-site benchmarks.
For WebVoyager~\citep{he2024webvoyager}, we retain 11 live websites and exclude Allrecipes, Booking, Google Flights, and Google Search because site changes or reCAPTCHA make their tasks invalid. 
Online-Mind2Web~\citep{onlinemind2web} adds 300 tasks across more than 100 live websites. 
We mainly use WebArena for controlled ablations, while the other two test live-site and cross-site skill transfer.

\paragraph{Models \& Baselines.}
We focus on two strong multimodal LLMs: GPT-5~\citep{gpt5} and Qwen-3.5-122B-A10B-Int4~\citep{qwen3.5} because skills are often used in the context of advanced models in the community. We also supplement results for Qwen3.6-35B-A3B-FP8 and Qwen3.5-9B to broaden the comparison though pilot experiments showed that sub-30B models could not reliably follow skill schemas.
We compare against (1) \emph{Vanilla}, a ReAct~\citep{yao2023reactsynergizingreasoningacting} agent without skills; (2) \emph{MAP} (Multi-Action Prediction), which generates up to 3 browser actions per step;  (3) SkillWeaver~\citep{skillweaver}; and (4) WALT~\citep{walt}. Note that skill acquisition is part of each method: SkillWeaver explores websites, WALT reverse-engineers site functions, and \ours uses synthetic trajectories. To isolate skill organization and deployment effectiveness from this source difference, we also re-deploy SkillWeaver and WALT skills in our framework (\emph{\ours + SkillWeaver/WALT}).

\paragraph{Implementation.}
All methods use a 30 interaction step budget. Skills are extracted by SynthAgent~\citep{wang2025adaptingwebagentssynthetic} on synthetic tasks for each benchmark website.
At inference, we retrieve up to 20 candidate skills per page via skill graph. The agent framework is implemented with browser-use~\citep{browser_use_github}. 
For benchmark evaluation, we follow the default settings by each benchmark and use GPT-4.1 if required for LLM judge.
More details are in Appendix~\ref{app:impl}.

\begin{table*}[t]
  \centering
  \small
  \setlength{\tabcolsep}{4pt}
  \renewcommand{\arraystretch}{1.0}
  \begin{tabular}{@{}llccccc|c@{}}
    \toprule
    \textbf{Model} & \textbf{Method} & \textbf{Shop.} & \textbf{CMS} & \textbf{Reddit} & \textbf{GitLab} & \textbf{Map} & \textbf{Overall} \\
    \midrule
    \multirow{8}{*}{GPT-5}
    & Vanilla              & 50.0 & 65.7 & 79.0 & 63.3 & 50.0 & 59.7 \\
    & Vanilla + MAP                    & 52.3 & 80.0 & 84.2 & 60.0 & 46.2 & 63.0 \\
    \cmidrule(l){2-8}
    & SkillWeaver~\citep{skillweaver}            & 28.3 & 42.9 & 30.2 & 44.4 & 41.3 & 37.4 \\
    & WALT~\citep{walt}                   & 30.9 & 47.4 & 57.7 & 51.8 & 28.1 & 42.9 \\
    \cmidrule(l){2-8}
    & \ours + SkillWeaver    & 56.8 & 57.1 & 68.4 & 66.7 & 53.9 & 59.7 \\
    & \ours + WALT           & 52.3 & 77.1 & 84.2 & 56.7 & 50.0 & 62.3 \\
    & \ours (Grounded)       & 65.9 & 65.7 & 100.0 & 70.0 & 57.7 & \textbf{69.5} \\
    & \ours (Guided)         & 59.1 & 80.0 & 89.5 & 63.3 & 61.5 & \underline{68.8} \\
    \midrule
    \midrule
    \multirow{8}{*}{Qwen}
    & Vanilla              & 40.9 & 48.6 & 57.9 & 43.3 & 42.3 & 45.5 \\
    & Vanilla + MAP                    & 38.6 & 47.1 & 73.7 & 43.3 & 46.2 & 47.1 \\
    \cmidrule(l){2-8}
    & SkillWeaver~\citep{skillweaver}            & 37.8 & 38.7 & 66.7 & 40.0 & 46.2 & 43.8 \\
    & WALT~\citep{walt}                   & 31.8 & 45.7 & 57.9 & 50.0 & 46.2 & 44.2 \\
    \cmidrule(l){2-8}
    & \ours + SkillWeaver    & 36.4 & 42.9 & 63.2 & 53.3 & 30.8 & 43.5 \\
    & \ours + WALT           & 31.8 & 45.7 & 57.9 & 50.0 & 46.2 & 44.2 \\
    & \ours (Grounded)       & 47.7 & 60.0 & 47.4 & 46.7 & 38.5 & \underline{48.7} \\
    & \ours (Guided)         & 47.7 & 57.1 & 63.2 & 50.0 & 57.7 & \textbf{53.9} \\
    \bottomrule
  \end{tabular}
  \caption{Results of GPT-5 and Qwen-3.5-122B on WebArena. \ours + SkillWeaver/WALT deploys the respective method's within our framework. Best and second-best overall results are \textbf{bolded} and \underline{underlined}.}
  \label{tab:main_results}
\end{table*}

\subsection{Main Results}
\label{sec:results}

\paragraph{WebArena.}
Table~\ref{tab:main_results} present results on WebArena.
We first find that both deployment modes of \ours outperform all baselines on the overall metric for both backbones, confirming that executable skills reduce planning errors and improve task completion.
SkillWeaver and WALT in their original implementations are inferior to MAP. However, when their skills are re-deployed inside \ours, performance recovers and even surpasses MAP. This isolates the contribution of the agent harness from the skill quality and ensures best-effort comparison.
GPT-5 slightly favors grounded mode, whereas Qwen3.5-122B favors guided mode. Manual inspection confirms that Qwen is less reliable at recovering from execution faults, and guided mode helps by giving agent full autonomy to adapt. We further study the mode selection preference later within this section.
\begin{table*}[t]
  \centering
  \small
  \setlength{\tabcolsep}{4pt}
  \resizebox{\textwidth}{!}{
  \begin{tabular}{@{}lccccccccccc|c@{}}
    \toprule
    \textbf{Method} & \textbf{Amzn} & \textbf{Apple} & \textbf{ArXiv} & \textbf{BBC} & \textbf{Camb.} & \textbf{Cour.} & \textbf{ESPN} & \textbf{GitHub} & \textbf{GMap} & \textbf{HF} & \textbf{Wolf.} & \textbf{Overall} \\
    \midrule
    Vanilla              & 90.2 & 65.1 & 76.7 & 54.8 & 74.4 & 73.8 & 59.1 & 85.4 & 73.2 & 74.4 & 65.2 & 71.9 \\
    Vanilla + MAP                  & 90.2 & 72.1 & 81.4 & 35.7 & 81.4 & 69.1 & 56.8 & 92.7 & 80.5 & 81.4 & 78.3 & 74.4 \\
    \midrule
    \ours (Grounded)     & 95.1 & 74.4 & 86.1 & 81.0 & 90.7 & 83.3 & 79.6 & 95.1 & 90.2 & 83.7 & 89.1 & \textbf{86.1} \\
    \ours (Guided)       & 82.9 & 74.4 & 81.4 & 71.4 & 90.7 & 73.8 & 79.6 & 95.1 & 95.1 & 74.4 & 91.3 & 82.7 \\
    \ours (Guided + WA)  & 97.6 & 79.1 & 76.7 & 71.4 & 90.7 & 78.6 & 93.2 & 87.8 & 92.7 & 83.7 & 84.8 & \underline{85.1} \\
    \bottomrule
  \end{tabular}
  }
  \caption{Results on WebVoyager with GPT-5. Guided + WA uses skills extracted from WebArena.}
  \label{tab:webvoyager}
\end{table*}

\begin{table}[t]
    \centering
    \footnotesize
    \setlength{\tabcolsep}{3.6pt}
    \begin{tabular}{@{}lccc@{}}
      \toprule
      \diagbox{\textbf{Method}}{\textbf{Judge}} & \textbf{GPT-4.1} & \textbf{GPT-5.1} & \textbf{WebJudge} \\
      \midrule
      Vanilla & 59.00 & 53.33 & 61.33 \\
      MAP & 60.33 & 57.33 & 68.67 \\
      \midrule
      Grounded & \textbf{68.67} & \textbf{67.00} & \textbf{72.00} \\
      Guided & 67.33 & 63.33 & 70.00 \\
      \midrule
      Guided + WA & 67.33 & 66.00 & 68.00 \\
      \bottomrule
    \end{tabular}
    \caption{Result on Online-Mind2Web with GPT-5 with three different judge models.}
    \label{tab:onlinemind2web_results}
\end{table}

\paragraph{WebVoyager and Online-Mind2Web.}
Table~\ref{tab:webvoyager} and Table~\ref{tab:onlinemind2web_results} present results on WebVoyager and Online-Mind2Web respectively.
For these two live websites benchmarks, we mainly compare against Vanilla and MAP methods, because SkillWeaver and WALT do not release skills for live websites and their acquisition pipelines rely on heavy exploration or website reverse-engineering that is not feasible for online services.
Grounded mode reaches 86.1\%, improving over Vanilla by 14.2 points and MAP by 11.7 on WebVoyager.
For Online-Mind2Web, grounded mode is consistently the best method under all three judges. The guided mode also improves over MAP by 4.8 points. 

\paragraph{Skill Transfer across Websites.}
Another advantage for \ours is that guided mode allows skill transfer across sites, while prior code-based methods rely on fixed action scripts that are tightly coupled with specific websites and cannot transfer. We test it with only WebArena skills (Guided + WA) and it reaches 85.1\% without target-site programs, a 10.7 point gain over MAP on WebVoyager, while it has 8.1\% relative improvement over MAP on Online-Mind2Web. 
The transfer works precisely because step-level guidance survives website change while fixed action scripts do not, directly addressing the concern that URL-based retrieval and grounded scripts cannot generalize across sites. 

\paragraph{Deployment Mode Preference.}
A central value of \ours is that a skill does not commit the agent to a single execution interface. The same learned procedure can be invoked as an executable tool or read as a step-by-step plan, allowing each backbone to use procedural knowledge in the form that best matches its capabilities. Table~\ref{tab:main_results} shows that GPT-5 favors grounded mode, whereas Qwen3.5-122B performs better with guided mode. The comparison in Table~\ref{tab:deployment_modes} extends this pattern: Qwen3.6-35B benefits from grounded execution, while Qwen3.5-9B benefits from guided use. These results point to a general design principle for agent skills: reusable knowledge and execution control should be decoupled. Models that can carry out a complete procedure as a tool gain from grounded execution; models that benefit from reasoning over intermediate steps can use the same skill as guidance while retaining control of their actions.

\begin{table}[t]
  \centering
  \footnotesize
  \resizebox{\columnwidth}{!}{%
  \begin{tabular}{@{}lcc|cc@{}}
    \toprule
    & \multicolumn{2}{c}{\textbf{WebArena}} & \multicolumn{2}{c}{\textbf{WebVoyager}} \\
    \cmidrule(lr){2-3}\cmidrule(l){4-5}
    \textbf{Backbone} & \textbf{Grounded} & \textbf{Guided} & \textbf{Grounded} & \textbf{Guided} \\
    \midrule
    Qwen3.6-35B-A3B & \textbf{59.09} & 57.79 & \textbf{89.98} & 85.93 \\
    Qwen3.5-9B & 44.16 & \textbf{49.35} & 71.86 & \textbf{84.86} \\
    \bottomrule
  \end{tabular}%
  }
  \caption{Skill deployment mode study with broader open-source models.}
  \label{tab:deployment_modes}
\end{table}

\begin{table*}[t]
  \centering
  \small
  \resizebox{0.95\textwidth}{!}{%
  \begin{tabular}{@{}lcccccccccccc@{}}
    \toprule
    & \multicolumn{6}{c}{Avg. Steps $\downarrow$} & \multicolumn{6}{c}{Skill Usage \%} \\
    \cmidrule(lr){2-7} \cmidrule(l){8-13}
    & & & & & & & \multicolumn{2}{c}{All} & \multicolumn{2}{c}{Succ.} & \multicolumn{2}{c}{Fail.} \\
    \cmidrule(lr){8-9} \cmidrule(lr){10-11} \cmidrule(l){12-13}
    \textbf{Method} & \textbf{Shop.} & \textbf{CMS} & \textbf{Red.} & \textbf{Git.} & \textbf{Map} & \textbf{All} & \textbf{IR} & \textbf{UR} & \textbf{IR} & \textbf{UR} & \textbf{IR} & \textbf{UR} \\
    \midrule
    Vanilla              & 7.8  & 11.9 & 7.5  & 11.6 & 13.3 & 10.4 & --   & --   & --   & --   & --   & --   \\
    Vanilla + MAP                  & 6.1  & 11.1 & 13.1 & 11.8 & 8.4  & 9.8  & --   & --   & --   & --   & --   & --   \\
    \midrule
    SkillWeaver          & 6.8  & 10.2 & 7.4  & 11.7 & 11.4 & 9.4  & 6.6  & 37.7 & 7.5  & 39.1 & 5.5  & 35.5 \\
    WALT                 & 6.5  & 11.0 & 7.3  & 11.5 & 10.0 & \textbf{9.2}  & 9.3  & 33.1 & 9.3  & 32.3 & 9.3  & 34.5 \\
    \midrule
    \ours (Grounded)     & 6.1  & 10.5 & 6.2  & 13.0 & 11.1 & \underline{9.3}  & \textbf{16.5} & \textbf{70.8} & 19.5 & 73.8 & 12.4 & 63.8 \\
    \ours (Guided)       & 7.5  & 11.5 & 8.6  & 14.5 & 11.1 & 10.5 & \underline{10.1} & \underline{68.8} & 11.4 & 68.9 & 8.3  & 68.8 \\
    \bottomrule
  \end{tabular}
  }
  \caption{Efficiency and skill usage analysis on WebArena with GPT-5. IR (Invocation Rate): percentage of agent steps that invoke a skill. UR (Usage Rate): percentage of tasks where at least one skill is invoked. Succ. and Fail. denote successful and failed tasks, respectively. }
  \label{tab:efficiency}

\end{table*}

\paragraph{Efficiency \& Skill Usage.}
Table~\ref{tab:efficiency} studies step efficiency and skill usage. 
\ours achieves substantially higher skill adoption than SkillWeaver and WALT: grounded mode reaches Usage Rate (UR) 70.8\% and Invocation Rate (IR) 16.5\%, roughly doubling WALT and SkillWeaver.
This higher adoption reflects not only the usefulness of our extracted skills but also the effectiveness of our skill organization.
Splitting by outcome, grounded mode shows notably higher IR and UR on successful tasks, suggesting that effective skill invocation contributes to task completion.
In contrast, Guided mode shows nearly identical UR across success and failure (68.9\% vs.\ 68.8\%), indicating that guided skills are uniformly available but success depends on the agent's own execution quality rather than skill availability.
For step efficiency, Grounded mode achieves 9.3 average steps with higher success rate, fewer steps than Vanilla.
Guided mode uses slightly more steps as the agent needs to take action by action, trading efficiency for adaptability.

\section{Analysis}

\begin{table}[t]
  \centering
  \footnotesize
  \setlength{\tabcolsep}{3pt}
  \begin{tabular}{@{}lccccc@{}}
    \toprule
    Method & \#Sk. & \#Op. & Sk./Nd. & SR & Util. \\
    \midrule
    SkillWeaver & 87.6 & 1.6 & -- & 84.1 & 8.2 \\
    WALT & 8.2 & 4.5 & -- & 67.2 & 22.0 \\
    \midrule
    \ours & 118.2 & 3.8 & 4.2 & 77.1 & 12.9 \\
    \ours$^\dagger$ & 67.6 & 3.5 & 3.5 & 85.0 & 27.8 \\
    \bottomrule
  \end{tabular}
  \caption{Skill quality and coverage. $^\dagger$: skills from test-set trajectories. \#Sk.: skills/site; \#Op.: write actions/skill; Sk./Nd.: skills/page node; SR: execution success (\%); Util.: skills invoked (\%).}
  \label{tab:skill_quality}
\end{table}

\begin{figure*}[t]
    \centering
    \includegraphics[width=1.0\textwidth]{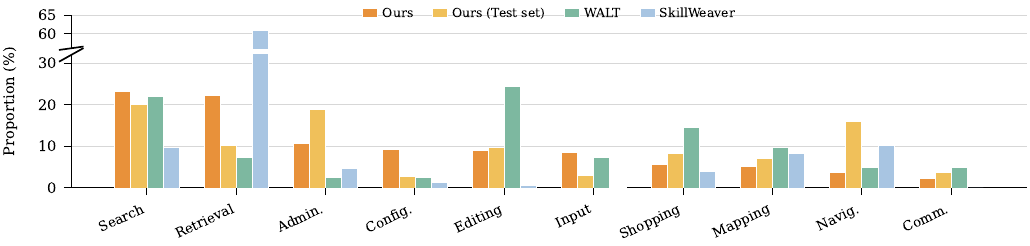}
    \caption{Skill category distribution. \ours covers all ten categories; SkillWeaver concentrates over 60\% in Retrieval, and WALT has 41 skills in total.}
    \label{fig:skill_category}
\end{figure*}

\begin{figure*}[t]
    \centering
    \includegraphics[width=1.0\textwidth]{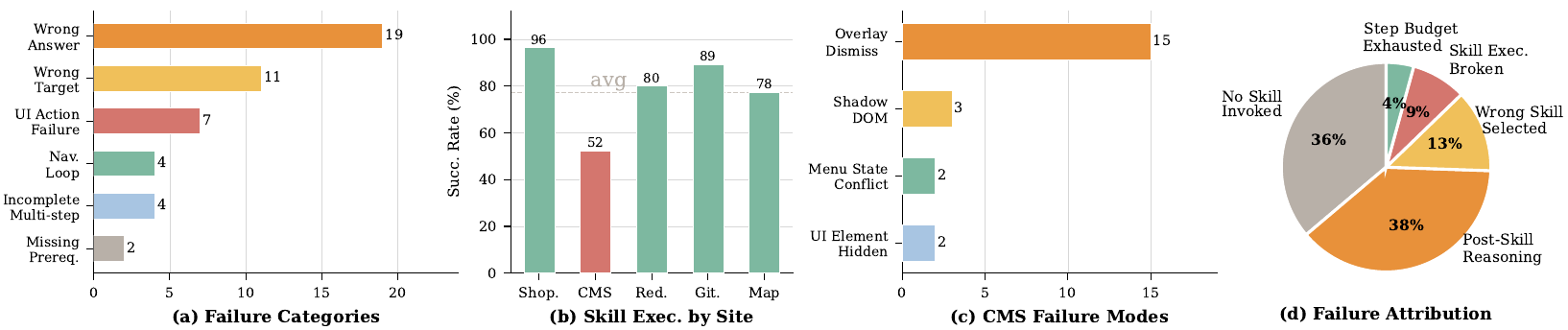}
    \caption{Failure analysis of \ours (grounded mode) on WebArena. (a) Failure categories for failed tasks from trajectory inspection. (b) Per-site skill execution success rate. (c) Root causes of CMS skill execution failures. (d) Failure attribution by skill usage role.}
    \label{fig:failure_analysis}
\end{figure*}

\subsection{Skill Quality and Coverage}

Table~\ref{tab:skill_quality} and Figure~\ref{fig:skill_category} compare skill libraries across methods.
\ours builds the largest library (100+ skills per site) from low-cost synthetic trajectories, spread across all ten functional categories with no single category exceeding 24\%. 
In contrast, SkillWeaver concentrates over 60\% in retrieval (data-extraction routines), and WALT provides only 41 skills in total, resulting in sparse coverage across most categories.
The category distribution of our synthetic skills closely matches the test-set oracle variant, indicating that our extraction captures the natural distribution of reusable skills without access to evaluation data.
Our skills are also more procedural, averaging 3.8 operations per skill compared to SkillWeaver's 1.6, since many of SkillWeaver's skills are value-extraction routines that only read page content.
Beyond library size, our skill graph keeps each page node compact (4.2 skills per node), which helps reduce retrieval noise. 
Utilization (Util.) highlights the difference between breadth and effective coverage: although \ours has $10\times$ more skills than WALT, its 12.9\% utilization corresponds to $\sim$15 unique skills invoked per site, while WALT's 22.0\% maps to only about 1.8 unique skills per site, far short of the procedural coverage long-horizon tasks demand, consistent with the results in Table~\ref{tab:efficiency}.
For execution reliability, the execution success rate (SR) of our curated and verified skills reaches 77\%, close to the 85\% achieved by the test-set oracle variant.

\begin{table}[t]
  \centering
  \footnotesize
  \setlength{\tabcolsep}{5pt}
  \begin{tabular}{@{}c|ccc@{}}
    \toprule
    \textbf{Corpus Size} & \textbf{\# Skills} & \textbf{TSR} & \textbf{UR} \\
    \midrule
    25\% & 105 & 62.8 & 26.6 \\
    50\% & 222 & 66.2 & 39.0 \\
    75\% & 474 & 68.8 & 51.9 \\
    100\% & 591 & \textbf{69.5} & \textbf{70.8} \\
    \bottomrule
  \end{tabular}
  \caption{Skill scaling on WebArena by varying the corpus size for skill extraction. TSR: task success rate (\%); UR: tasks invoking at least one skill (\%).}
  \label{tab:library_scaling}
\end{table}

\begin{table}[t]
  \centering
  \footnotesize
  \setlength{\tabcolsep}{3.2pt}
  \begin{tabular}{@{}llcc@{}}
    \toprule
    \textbf{Corpus} & \textbf{Organization} &\textbf{\# Skills} & \textbf{TSR} \\
    \midrule
    Full & Skill Graph & 591 & \textbf{69.5} \\
    \midrule
    +Filtering & Skill Graph & 548 & 68.8 \\
    Full & DOM matching & 591 & 65.6 \\
    Full & Flat Library & 591 & 59.1 \\
    \bottomrule
  \end{tabular}
  \caption{Study on corpus filtering and skill retrieval on WebArena with GPT-5. Filtered corpus removes sources similar to test tasks. DOM matching retrieves skills based on DOM structure similarity, while Flat library searches the full library.}  
  \label{tab:retrieval_controls}
\end{table}

\subsection{Skill Scaling and Retrieval}
\label{sec:data_analysis}

Table~\ref{tab:library_scaling} traces how the size of the 2,500-trajectory corpus changes the verified skill library and its use. Expanding from 25\% of the corpus to the full set grows the library from 105 to 591 skills, raises task success from 62.8\% to 69.5\%, and increases UR from 26.6\% to 70.8\%. All three measures rise monotonically, but coverage continues to grow most strongly in the long tail: the final 25\% of sources adds 117 skills and raises UR by 18.9 points. These later skills support increasingly specialized workflows, broadening the set of tasks for which the agent can retrieve a relevant procedure.

Table~\ref{tab:retrieval_controls} studies the generality of the extracted procedures with a similarity-filtered corpus. We retrieve the closest WebArena test task for each synthetic source, classify workflow and goal similarity with GPT-5, and remove 196 near-matching sources before rebuilding the library. The filtered corpus retains 548 verified skills and reaches 68.8\% task success, close to the full corpus at 69.5\%. The extracted procedures therefore remain useful across varied task instructions that share reusable browser operations.
We also hold the full skill library fixed and validate how skill candidates are selected. A flat-library search reaches 59.1\%, DOM matching reaches 65.6\%, and the proposed URL-based skill graph reaches 69.5\%. DOM matching gains 6.5 points by conditioning retrieval on the current page structure; the graph adds another 3.9 points by first narrowing candidates to the page's functional context and then filtering them against visible elements. This layered retrieval keeps the candidate set compact while preserving the coverage of full library.

\subsection{Ablation Study}
\label{sec:ablation}

\begin{table*}[t]
  \centering
  \resizebox{0.87\textwidth}{!}{%
  \begin{tabular}{@{}lcccccccccccc@{}}
    \toprule
    & \multicolumn{6}{c}{\textbf{Task Success Rate (\%)}} & & \multicolumn{5}{c}{\textbf{Skill Usage (\%)}} \\
    \cmidrule(lr){2-7} \cmidrule(lr){9-13}
    & & & & & & & & \multicolumn{2}{c}{\textbf{All}} & \multicolumn{2}{c}{\textbf{Succ.}} & \textbf{Fail.} \\
    \cmidrule(lr){9-10} \cmidrule(lr){11-12} \cmidrule(lr){13-13}
    \textbf{Variant} & \textbf{Shop.} & \textbf{CMS} & \textbf{Red.} & \textbf{Git.} & \textbf{Map} & \textbf{All} & \textbf{Steps} & \textbf{IR} & \textbf{UR} & \textbf{IR} & \textbf{UR} & \textbf{UR} \\
    \midrule
    \ours (Grounded) & \textbf{65.9} & 65.7 & \textbf{100} & 70.0 & 57.7 & \textbf{69.5} & 9.3 & 16.5 & 70.8 & 19.5 & 73.8 & 63.8 \\
    ~~w/ Test-set Skills & 59.1 & \textbf{71.4} & 84.2 & \textbf{86.7} & 46.2 & 68.2 & \textbf{8.1} & \textbf{28.2} & \textbf{86.4} & \textbf{29.6} & \textbf{83.8} & \textbf{91.8} \\
    ~~w/ Mix Mode & 59.1 & 68.6 & 84.2 & 70.0 & \textbf{57.7} & 66.2 & 8.6 & 17.9 & 70.1 & 19.3 & 72.5 & 65.4 \\
    \midrule
    ~~w/o Skill Verification & 40.9 & 60.0 & 78.9 & 60.0 & 50.0 & 55.2 & 10.4 & 21.2 & 74.0 & 19.1 & 72.9 & 75.4 \\
    ~~w/o Skill Graph & 52.3 & 51.4 & 94.7 & 56.7 & 57.7 & 59.1 & 10.9 & 10.1 & 56.5 & 11.0 & 53.8 & 60.3 \\
    ~~w/o Step Guidance & 52.3 & 62.9 & 84.2 & 63.3 & 50.0 & 60.4 & 10.0 & 17.6 & 73.4 & 19.1 & 75.3 & 70.5 \\
    \bottomrule
  \end{tabular}%
  }
  \caption{Ablations on WebArena with GPT-5. IR: percentage of steps invoking a skill; UR: percentage of tasks invoking at least one skill.}
  \label{tab:ablation}
\end{table*}

\begin{table}[!htbp]
  \centering
  \footnotesize
  \setlength{\tabcolsep}{2.5pt}
  \begin{tabular}{@{}llccc@{}}
    \toprule
    \textbf{Backbone} & \textbf{Mode} & \textbf{TSR} & \textbf{UR} & \textbf{Gr. choices} \\
    \midrule
    \multirow{3}{*}{GPT-5} & Grounded & 69.5 & 70.8 & 100.0 \\
    & Guided & 68.8 & 68.8 & 0.0 \\
    & Mix & 66.2 & 70.1 & 97.1 \\
    \midrule
    \multirow{3}{*}{Qwen3.5-122B} & Grounded & 48.7 & 48.4 & 100.0 \\
    & Guided & 53.9 & 21.4 & 0.0 \\
    & Mix & 42.2 & 40.3 & 96.2 \\
    \bottomrule
  \end{tabular}
  \caption{Mix-mode behavior on WebArena. TSR/UR and grounded (Gr.) choices are percentages.}
  \label{tab:mix_behavior}
\end{table}

Table~\ref{tab:ablation} isolates each component, and the contributions align with the four axes of the grounding gap (Table~\ref{tab:cmp_baseline}). We have the following observations: 
(1) Broad experience produces strong procedural coverage. The synthetic library reaches 69.5 task success, comparable to 68.2 from an oracle library built with test-set trajectories, showing that diverse synthetic experience can recover the workflows needed at evaluation time. 
(2) Skill value comes from the whole lifecycle, not extraction alone. Verification makes the largest contribution, while contextual organization and step-level guidance each provide substantial gains. A useful skill must be reliable in execution, surfaced in the right context, and understandable to the agent when adaptation is needed. 
(3) The framework improves skills beyond those it learns itself. Organizing and deploying SkillWeaver and WALT skills with \ours raises success by 22.3 and 19.4 points, respectively, showing that our organization and deployment provide a reusable interface for existing skill collections. 
(4) Skill deployment should match how a model uses procedural knowledge. Each model exhibits a different preference between grounded and guided use, while mix mode largely converges to grounded execution. The useful flexibility is therefore at the model level: the same skill can be exposed as an executable tool or a readable plan according to the model's capabilities. 

\subsection{Failure Analysis}

We manually inspect all failed trajectories in Figure~\ref{fig:failure_analysis} and find:
(1) Failures fall into six categories, with wrong answer extraction as the dominant mode: the agent completes the workflow correctly but reports an incorrect final answer.
(2) Execution reliability varies sharply by site, from 96\% on Shopping down to 52\% on CMS, the most structurally complex environment in WebArena. 
(3) The dominant CMS execution failure, ``overlay dismiss'', is usually benign: these skills end with a click on a non-interactive element intended to dismiss a navigation overlay. The preceding steps succeed and 80\% of affected tasks still complete despite the trailing error, so strict per-step success can understate end-task utility.
(4) Failure attribution shows that post-skill reasoning (38\%) is the largest failure category: skills executed successfully but the agent failed at subsequent steps. No skill invoked (36\%) means the agent bypassed available skills entirely and failed using only native browser actions. 
Overall, most failures originate from agent-level decision making rather than skill design flaws, suggesting that further gains may be obtained by improving the reasoning and planning of LLM and the context management of agent harness.

\FloatBarrier
\section{Conclusion}

We introduce \ours, a skill-learning framework for autonomous web agents that bridges the grounding gap in existing skill formulations.
By making action programs with step-level natural language guidance visible and readable to the agent, \ours enables both efficient grounded execution and adaptable guided use, while skill extraction from synthetic trajectories and organization in a URL-based graph ensure a rich and retrievable skill library.
Across a series of benchmark tasks, \ours consistently outperforms strong baselines. 
Extensive analysis reveals a model-dependent preference for deployment mode: stronger models benefit from grounded execution, while weaker models benefit from guided use. 
Overall, \ours offers a practical approach for equipping web agents with reusable procedural knowledge through readable and executable skills.

\clearpage
\section*{Limitations} \label{sec:limitation}
While \ours advances skill learning for web agents, it has several limitations that may be further improved in future work:
\begin{itemize}[leftmargin=*]
  \item Our skill extraction relies on a corpus of synthetic agent trajectories, while this is viewed as a strength for its scalability and avoidance of test-data leakage, it also inherits the blind spots of the upstream synthesis. Sites whose interactions are poorly covered by synthetic trajectories will yield a sparser skill library. 
  \item The URL-based skill graph is primarily designed for in-site retrieval, which may raise concerns about limiting cross-site transfer. However, this limitation stems from the nature of code-based web skills rather than the retrieval mechanism itself: such skills, including ours, are tightly coupled to site-specific DOM structures and are therefore inherently difficult to transfer across sites. We introduce guided mode to partially relax this constraint by exposing skills as procedural guidance rather than fixed scripts.
\end{itemize}

\section*{Use of AI Assistants} \label{sec:ai_assistants}
We used coding agents (e.g., Claude Code) to assist with code implementation and debugging. And we also used AI assistants (e.g., ChatGPT) to help polish the writing of the paper. The authors have reviewed and edited the content as needed.

\bibliography{custom}

\clearpage
\newpage
\appendix
\section{Implementation Details}
\label{app:impl}

\subsection{Synthetic Trajectory Collection}
\label{app:synthagent}

Our skills are extracted from trajectories collected by SynthAgent~\citep{wang2025adaptingwebagentssynthetic}, which synthesizes diverse web tasks through categorized exploration of target websites.
We use 2,500 and 600 synthetic tasks for WebArena~\citep{webarena} and WebVoyager~\citep{he2024webvoyager}, respectively, covering a wide range of user intents and website functionalities.
Our extraction pipeline is designed to mine reusable action subsequences from both successful and failed trajectories, since a failed trajectory may still contain useful skills to finish a subtask before the failure point.
For Online-Mind2Web~\citep{onlinemind2web}, we generate independent tasks from site names and home URLs with SynthAgent.
Benchmark instructions, labels, outcomes, and execution trajectories are not supplied to task synthesis or skill extraction.

\subsection{Skill Extraction Details}
\label{app:extraction_details}

\paragraph{Trajectory Formatting.}
Each trajectory is converted to a structured textual representation before being sent to the LLM. For each step, we record the page URL, the agent's reasoning (thinking and next goal), and each action with its type, target element description (tag name, visible text, and key HTML attributes such as \texttt{id}, \texttt{name}, \texttt{aria-label}), and action parameters.
The same extraction prompt and settings are used for every benchmark. The output must follow the JSON schema in Table~\ref{tab:prompt_extraction_2}; each candidate contains 3--6 meaningful actions and guidance for every action.

\paragraph{Online Deduplication.}
When processing the trajectory corpus sequentially, each newly proposed skill is compared against the existing library using a three-level cascade:
(1) exact name deduplication to filter identical skill names;
(2) same-site near-duplicate filtering based on Jaccard similarity between tokens in skill function names;
(3) library-wide matching using a similarity score with OpenAI's text-embedding-3-small model.
For each candidate, the top-$k$ ($k{=}20$) most similar existing skills are included in the extraction prompt to help the LLM decide whether to add, update, or skip. The full extraction prompt is shown in Tables~\ref{tab:prompt_extraction_1} and~\ref{tab:prompt_extraction_2}.

\begin{table}[!htbp]
\begin{promptbox}{Skill Extraction Prompt (1/2): Task \& Deduplication}
You are a web automation expert. Analyze the trajectory and
extract reusable skill abstractions.

## Minimum Steps
Each skill MUST have at least {step_threshold} meaningful action
steps (click, input, send_keys, navigate, etc.).

## Duplicate & Overlap Detection (Quality Over Quantity)
Before creating any new skill, carefully check the existing
library below. Each skill has a similarity_score.
You MUST "skip" or "update" (NOT "new") if ANY of these:
1. Same user goal: An existing skill achieves the same outcome.
2. Large semantic overlap: >=70% overlapping action steps.
3. Differ only in final step(s).
[...]

## Existing Skill Library
{existing_skills_section}

## Trajectory
{trajectory_text}

## Your Task
1. Identify reusable action sequences ({step_threshold}-6
   actions) that could be abstracted into skills.
2. Skills should be generic (parameterized), atomic (single
   logical operation), and reusable across different tasks.
3. Check the existing library first. For each candidate skill:
   - similarity_score > 0.4 -> very likely overlaps
   - Same user goal -> "skip"
   - Better version in trajectory -> "update"
   - No similar existing skill -> "new"

## meta_url Rules
- start_url: The EXACT URL from the trajectory
- meta_url: A generalized URL pattern
  - Use * for variable parts: "gitlab/*/*/-/issues/*"
\end{promptbox}
\caption{Skill extraction prompt (1/2): task description, deduplication rules, and skill abstraction instructions. The prompt receives the formatted trajectory and existing skill library with similarity scores.}
\label{tab:prompt_extraction_1}
\end{table}

\begin{table}[!htbp]
\begin{promptbox}{Skill Extraction Prompt (2/2): Output Format}
## Output Format
{
  "extractions": [
    {
      "action": "new",
      "skill": {
        "name": "skill_name_in_snake_case",
        "description": "What this skill does",
        "start_url": "exact URL from trajectory",
        "meta_url": "generalized URL pattern",
        "parameters": [
          {"name": "p", "type": "str",
           "description": "...", "required": true}
        ],
        "action_steps": [
          {
            "guidance": "Why this action is needed",
            "action_type": "click|input|scroll|...",
            "element_ref": {
              "tag_name": "button|input|a|div|...",
              "text_content": "visible text",
              "attributes": {"id": "...", "class": "..."}
            },
            "params": {"text": "{{param_name}}"}
          }
        ]
      }
    },
    {"action": "skip", "existing_id": "skill_42",
     "reason": "Same as search_product"},
    {"action": "update", "existing_id": "skill_42",
     "reason": "Shorter: 3 steps vs 4",
     "skill": { ... complete updated skill ... }}
  ]
}

## Action Types Reference
1. click    - element_ref: REQUIRED
2. input    - element_ref: REQUIRED, params: {"text","clear"}
3. select_dropdown - element_ref: REQUIRED, params: {"text"}
4. scroll   - element_ref: null, params: {"direction","pages"}
5. send_keys - element_ref: null, params: {"keys"}
6. navigate - element_ref: null, params: {"url","new_tab"}
[...]

## Important Rules
- guidance is REQUIRED for every action step
- Use {{param_name}} syntax for parameterized values
- Even failed trajectories may contain useful action sequences
\end{promptbox}
\caption{Skill extraction prompt (2/2): output JSON format, action types reference, and important rules. The LLM outputs extraction decisions (\texttt{new}, \texttt{skip}, or \texttt{update}) with complete skill definitions.}
\label{tab:prompt_extraction_2}
\end{table}

\subsection{Skill Format}
\label{app:skill_format}

As shown in Table~\ref{tab:skill_example}, each executable skill is represented as a JSON object containing a semantic signature (name, description, typed parameters) and a sequence of action steps. Every action step includes a \texttt{guidance} field providing natural language guidance, an \texttt{action\_type}, an \texttt{element\_ref} describing the target UI element via tag name and HTML attributes, and step-specific \texttt{params}. Parameters use the \texttt{\{\{param\_name\}\}} syntax for value abstraction.

\begin{table}[!htbp]
\begin{promptbox}{Example of an Extracted Executable Skill (JSON)}
{
  "name": "search_products_from_homepage",
  "description": "From the shopping homepage, perform a product
    search using the main search box and submit it.",
  "meta_url": "shopping",
  "parameters": [
    {"name": "query", "type": "str",
     "description": "Product search query to enter.",
     "required": true}
  ],
  "action_steps": [
    {
      "guidance": "Focus the main site search input.",
      "action_type": "click",
      "element_ref": {
        "tag_name": "input",
        "attributes": {"id": "search", "name": "q",
          "placeholder": "Search entire store here..."}
      },
      "params": {}
    },
    {
      "guidance": "Type the search query, clearing existing text.",
      "action_type": "input",
      "element_ref": {
        "tag_name": "input",
        "attributes": {"id": "search", "name": "q"}
      },
      "params": {"text": "{{query}}", "clear": true}
    },
    {
      "guidance": "Submit the search by clicking the Search button.",
      "action_type": "click",
      "element_ref": {
        "tag_name": "button", "text_content": "Search",
        "attributes": {"type": "submit", "aria-label": "Search"}
      },
      "params": {}
    }
  ]
}
\end{promptbox}
\caption{Example of an extracted executable skill in JSON format. Each skill contains a semantic signature (name, description, typed parameters) and a sequence of action steps with element references and natural language step-level guidance.}
\label{tab:skill_example}
\end{table}

\subsection{Skill Graph and Retrieval}
\label{app:graph_details}

The skill graph uses generalized URL patterns (\texttt{meta\_url}) as node identifiers, where variable path segments are replaced with wildcards (e.g., \texttt{gitlab/*/*/-/issues/*}).
At inference time, URL matching follows a priority-based aggregation:
(1) exact match on the normalized URL (highest priority);
(2) wildcard pattern matching via \texttt{fnmatch} on all graph nodes, ranked by specificity (number of non-wildcard path segments).
Skills from all matching nodes are merged, with more specific matches taking priority and duplicates (by skill name) removed.
Up to 20 skills are surfaced per page. During runtime, we also use element-based heuristics to filter skills whose target elements are not present on the current page, improving relevance and reducing noise for the agent.

\subsection{Skill Deployment Prompts}
\label{app:deployment_details}

In grounded mode (\S\ref{sec:deployment}), skills are registered as callable tools (prefixed with \texttt{fg\_} to differ from native browser actions like \texttt{click}) in the agent's action space. The agent system prompt includes action rules explaining the availability and recommended usage of these pre-built skills (Table~\ref{tab:prompt_grounded_rules}). Each registered tool carries a description derived from the skill's metadata, including action step guidance and parameter specifications (Table~\ref{tab:prompt_grounded_example}).

\begin{table}[!htbp]
\begin{promptbox}{Grounded Mode: Action Rules}
<action_rules>
- In some pages, you are equipped with pre-built skills (fg_*).
  Each skill is a sequence of low-level actions (click, input,
  scroll, wait, etc.) that will be sequentially executed when
  you call the skill.
- Prefer fg_* skills when they match your current goal because
  they are efficient and well-tested.
- If an fg_* skill fails or does not meet your expectations,
  do not repeat the same failing fg_* call; continue using
  native browser actions to achieve your goal.
</action_rules>

<reasoning_rules>
[...]
- If current page provides pre-built skills (fg_*), you need
  to provide reasoning about whether to use them or not.
</reasoning_rules>
\end{promptbox}
\caption{Grounded mode action rules appended to the agent system prompt, instructing the agent on how to use pre-built skill tools (\texttt{fg\_*}).}
\label{tab:prompt_grounded_rules}
\end{table}

\begin{table}[!htbp]
\begin{promptbox}{Grounded Mode: Example Registered Skill Tool}
Tool name: fg_search_products_from_homepage
Description: From the shopping homepage, perform a product
search using the main search box and submit it.

Action Steps:
  Step 1: Focus the main site search input. ->
    click <input id="search" placeholder="Search entire
    store here...">
  Step 2: Type the search query into the main search field.
    -> input <input id="search"> with text="{{query}}"
  Step 3: Submit the search by clicking the Search button.
    -> click <button type="submit" aria-label="Search">

Parameters (pass as kwargs_json):
  - query: str (required) - Product search query to enter.
\end{promptbox}
\caption{Example of a registered skill tool in grounded mode. The tool description includes the skill's action step guidance and parameter specifications.}
\label{tab:prompt_grounded_example}
\end{table}

In guided mode, skills are surfaced as step-by-step workflow instructions rather than auto-executed tools. The agent system prompt includes a skill mode section explaining how to activate and follow skills (Table~\ref{tab:prompt_guided_sys}). At runtime, available skills are listed via an \texttt{<available\_skills>} block. Upon activation, step-level guidance is injected via \texttt{<activated\_skill\_guidance>} (Table~\ref{tab:prompt_guided_runtime}).

\begin{table}[!htbp]
\begin{promptbox}{Guided Mode: System Prompt}
## SKILL MODE

You have access to a library of Skills -- pre-defined
step-by-step workflow guides for common tasks on web pages.

### How Skills Work:
- Each step, you'll see <available_skills> listing skills
- Call use_skill(skill_name="...") to activate a skill
- On your NEXT step, <activated_skill_guidance> will appear
  with detailed step-by-step instructions
- Follow the guidance using your normal browser actions
- Skills are guides, not automatic executors -- YOU perform
  each step
- Call clear_skill() when finished

### When to Use Skills:
- When an available skill matches your current sub-goal
- Skills save you from figuring out the UI workflow

### When NOT to Use Skills:
- When no skill matches your goal
- When the task is simple enough

### After Activating a Skill:
1. Read the step-by-step guidance carefully
2. Observe the page to identify the elements described
3. Execute each step using your browser actions
4. If a step fails, adapt and continue
5. Call clear_skill() when done
6. IMPORTANT: skill is a general guide, not a strict script;
   use your judgment to adapt as needed.
\end{promptbox}
\caption{Guided mode system prompt appended to the agent, explaining how to activate and follow skill guidance using native browser actions.}
\label{tab:prompt_guided_sys}
\end{table}

\begin{table}[!htbp]
\begin{promptbox}{Guided Mode: Runtime Skill Injection}
--- Injected into agent's input message each step ---

<available_skills>
Skills are pre-defined step-by-step workflow guides.
Call use_skill(skill_name="...") to activate a skill.
Call clear_skill() when finished.

1. search_products_from_homepage: Perform a product search
   using the main search box.
2. navigate_to_category: Navigate to a specific product
   category from the navigation menu.
3. sort_search_results: Sort search results by a criterion.
[...]
</available_skills>

--- After calling use_skill("search_products_...") ---

<activated_skill_guidance>
Skill: "search_products_from_homepage"
Description: Perform a product search using the main search
box and submit it with the Search button.

Follow these steps using your browser actions:

Step 1: Focus the main site search input.
  Example: click on <input> placeholder="Search entire store
  here..." id="search"
Step 2: Type the search query, clearing existing text.
  Example: input "{{query}}" into <input> id="search"
Step 3: Submit the search by clicking the Search button.
  Example: click on <button> text="Search" type="submit"

IMPORTANT:
- Use native browser actions to execute each step
- If a step fails, adapt and continue
- Call clear_skill() when finished
</activated_skill_guidance>
\end{promptbox}
\caption{Guided mode runtime injection. Available skills are listed via \texttt{<available\_skills>}; upon activation, step-by-step guidance with action examples is injected via \texttt{<activated\_skill\_guidance>}.}
\label{tab:prompt_guided_runtime}
\end{table}

\subsection{Web Agent Evaluation}
\label{app:reliability}
Evaluating web agents can be challenging due to the time-consuming nature of browser actions and costly live site interactions.
Across the web agent literature~\citep{walt,skillweaver,asi}, it is common practice to evaluate on a single benchmark such as WebArena~\citep{webarena} because it provides high-quality verification and reproducible self-hosted environments, allowing for more reliable attribution of performance differences to the agent design rather than environmental noise.
To further test the generalization of \ours, we also evaluate on WebVoyager~\citep{he2024webvoyager}, which includes 10+ live websites. We mitigate reliability concerns by manually inspecting the trajectories to exclude unstable websites due to reCAPTCHA or task inaccessibility issues.
For Online-Mind2Web, we use GPT-4.1, GPT-5.1 and WebJudge-7B to evaluate the 300 tasks with the benchmark's key-point extraction, screenshot selection, and final-decision pipeline with a threshold of 3 for task success.

\section{Case Study}
\label{app:casestudy}

We present two examples from WebArena using GPT-5 that illustrate how \ours skills assist the agent in both grounded and guided modes. For each case, we show the task description, selected trajectory steps with browser screenshots, and the agent's actions. Steps where a skill is invoked or activated are highlighted with a green background.

\subsection{Grounded Mode}

\begin{figure}[t]
  \centering
  \includegraphics[width=\columnwidth]{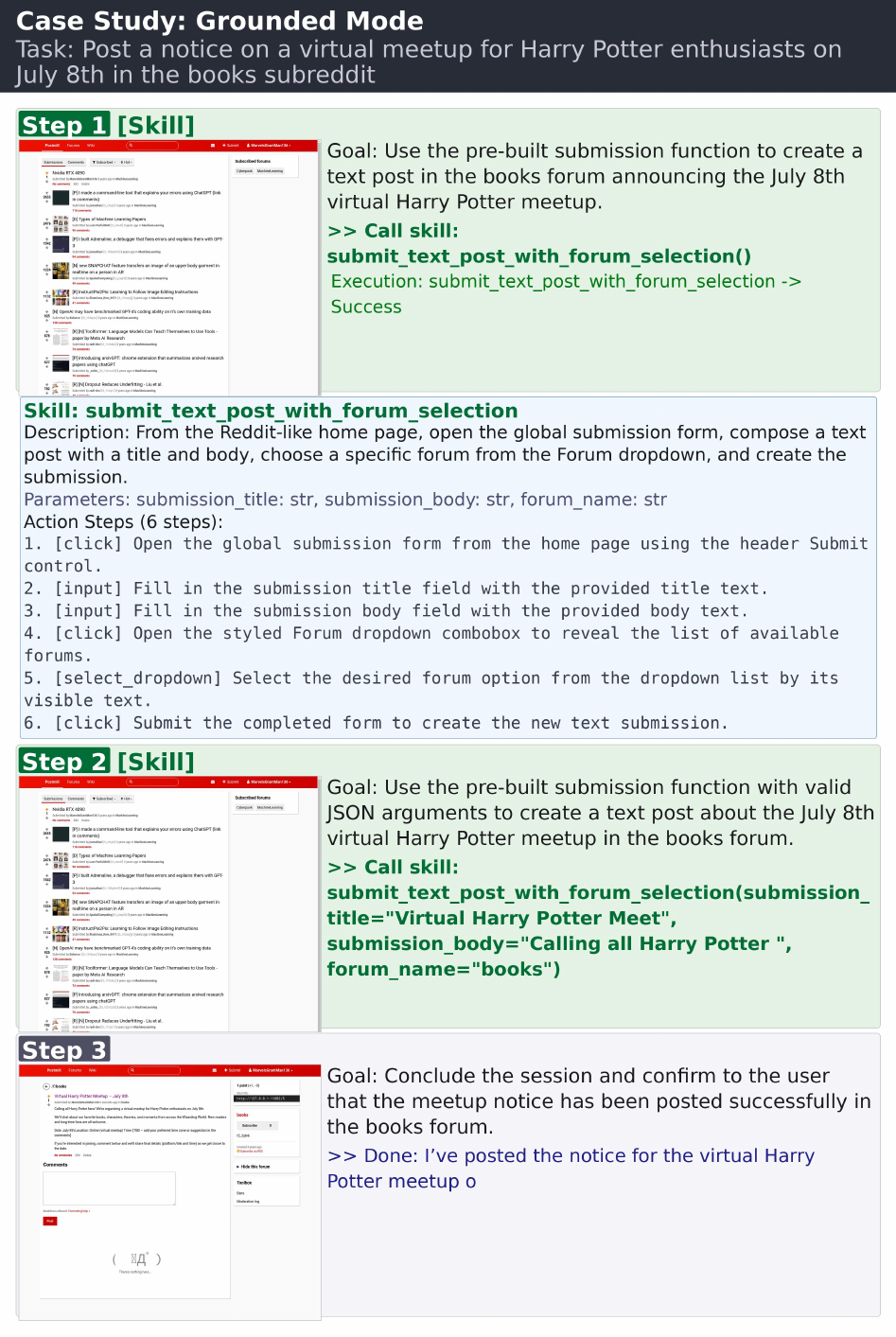}
  \caption{Grounded mode case study: the agent completes a Reddit forum posting task in 3 steps by invoking a skill that auto-executes the multi-step post submission workflow.}
  \label{fig:case_grounded}
\end{figure}

Figure~\ref{fig:case_grounded} shows the agent completing a Reddit forum posting task in only 3 steps using grounded mode.
The task requires creating a post about a Harry Potter virtual meetup in the books subreddit.
In Step~0, the agent identifies the available skill \texttt{submit\_text\_post\_with\_forum\_selection} and invokes it with the post title, body, and target forum as parameters. The first attempt fails due to a JSON formatting error. In Step~1, the agent retries the same skill with corrected arguments, and the runtime automatically executes the entire multi-step workflow: navigating to the submission form, selecting the ``books'' forum, filling in the title and body, and submitting. In Step~2, the agent confirms the post is visible and calls \texttt{done}.
Without the skill, this task would require the agent to independently navigate to the correct forum, find the submission form, fill in multiple fields, and submit, a sequence that typically takes 8--10 steps.

\subsection{Guided Mode}

\begin{figure}[t]
  \centering
  \includegraphics[width=\columnwidth]{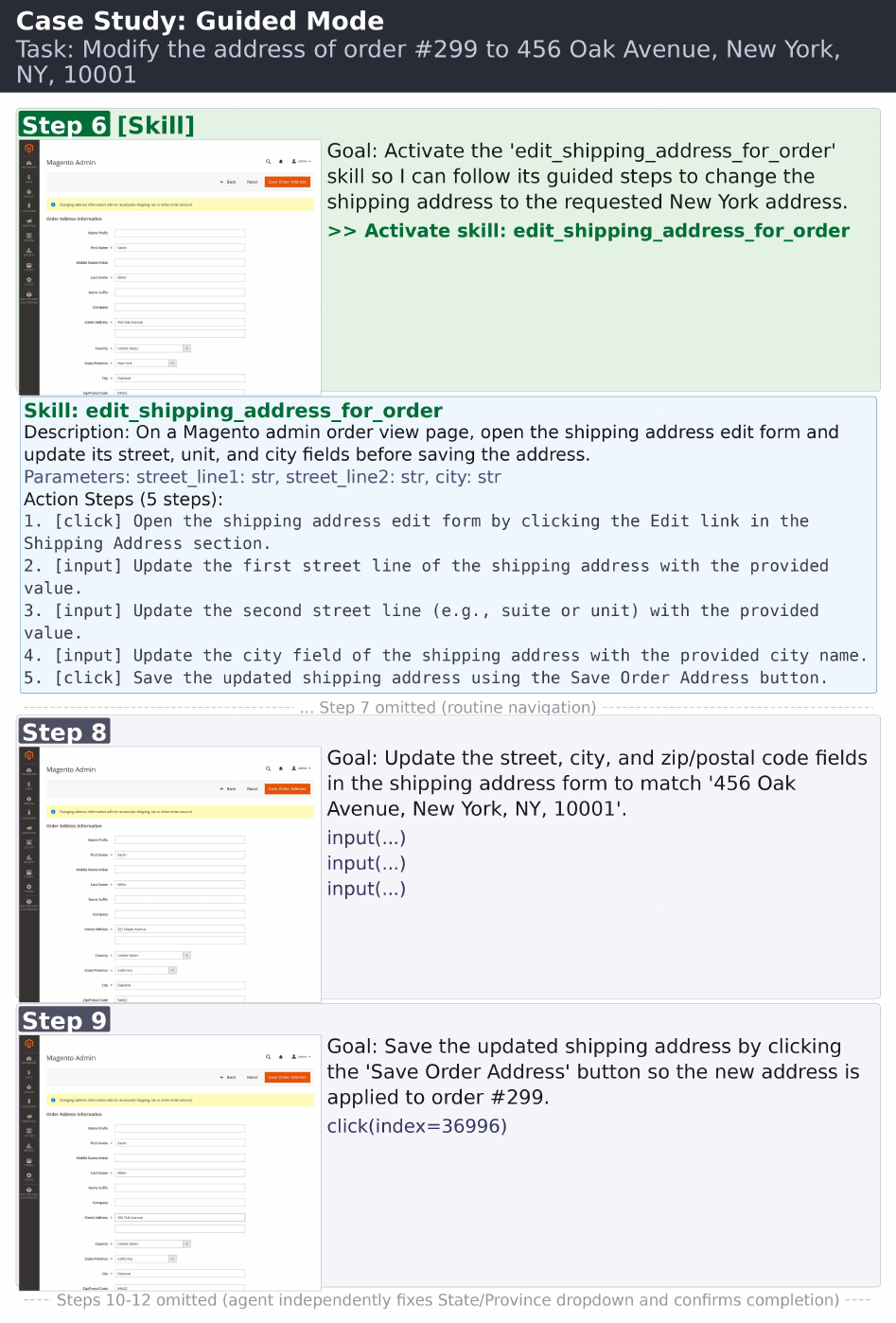}
  \caption{Guided mode case study: the agent modifies an order's shipping address by activating the skill for step-by-step guidance, while independently adapting when the skill does not cover all required fields.}
  \label{fig:case_guided}
\end{figure}

Figure~\ref{fig:case_guided} shows the agent modifying an order's shipping address in the CMS admin panel using guided mode (earlier navigation steps via the \texttt{open\_orders\_grid\_from\_any\_admin\_page} skill are omitted for brevity).
In Step~6, the agent activates the skill \texttt{edit\_shipping\_address\_for\_order}, which provides 5-step guidance for editing the address form. Following the guidance, the agent fills in the street, city, and zip code fields (Step~8) and saves the form (Step~9).
In the omitted final steps, the agent discovers that the State/Province field was not updated because the skill guidance did not cover dropdown fields. The agent independently re-opens the edit form and selects ``New York'' from the State/Province dropdown, demonstrating skill adaptation: the ability to go beyond the skill's step-by-step guidance when the actual page state changes.

\end{document}